\documentclass[preprint,12pt]{elsarticle}

\usepackage{amssymb}
\usepackage{amsmath}
\journal{XXX}

\begin{document}

\begin{frontmatter}



\title{Merging plans with incomplete knowledge about actions and goals through an agent-based reputation system}


\author{Javier Carbo, Miguel A. Patricio, Jose M. Molina}

\address{Av Universidad, Campus Colmenarejo, Univ. Carlos III de Madrid}

\begin{abstract}
Managing transition plans is one of the major problems of people with cognitive disabilities. Therefore, finding an automated way to generate such plans would be a helpful tool for this community. In this paper we have specifically proposed and compared different alternative ways to merge plans formed by sequences of actions of unknown similarities between goals and actions executed by several operator agents which cooperate between them applying such actions over some passive elements (node agents) that require additional executions of another plan after some time of use. Such ignorance of the similarities between plan actions and goals would justify the use of a distributed recommendation system that would provide an useful plan to be applied for a certain goal to a given operator agent, generated from the known results of previous executions of different plans by other operator agents. Here we provide the general framework of execution (agent system), and the different merging algorithms applied to this problem. The proposed agent system would act as an useful cognitive assistant for people with intellectual disabilities such as autism. 
\end{abstract}

\begin{keyword}
Recommendation \sep Reputation \sep Trust \sep Agents \sep Planning



\end{keyword}

\end{frontmatter}

\section{Introduction}

\subsection{General context and motivation}
Transition planning is remarkable issue to be solved by people with cognitive disabilities such as autism, it severily handicaps their daily life \cite{ShogrenPlotner2012}. The use of cognitive artifacts to solve this problem has been addressed before as it is established in \cite{Ronmark2014}. Our contribution to this issue is a domain-independent completely autonomous way to generate such plans that would be of much help for this community. 

The term multi-agent planning have been used in literature applied to two different kinds of problems: it has been interpreted as the problem to find plans for a group of agents \cite{EphratiRosenchein1993} \cite{Rosenschein1982}. In such approach, autonomous agents intend to coordinate their actions to satisfy the different goals of each agent \cite{Georgeff1983} \cite{MuscettolaSmith1987}. The problem that may arise in this problem definition, is that agents as they pursue their particular goals, may adopt a non cooperative attitude if they do not have incentives to do it. 

On the other hand, when agents share a common collective goal, multi-agent planning is seen as a "`divide et impera"' problem, where such common goal (typically complex) is split in subproblems among the agents that, afterwards have to be merged into a unique solution. Each agent solves a subproblem \cite{WilkinsMyers1998} \cite{EphratiRosenchein1993} that has be combined with the solutions of the other subproblems into an useful joint solution. In some cases, there is no partition of the problem at all, and agents just apply different solving algorithms to the same problem (as in scheduling for transportation systems \cite{Fischeretal1995} and when different planners are applied to converge faster to a solution). So merging plans problem have also been addressed by agent systems without a previous partition (and assignment) of a shared problem into subproblems. 

These last merging plans problems without previous partition are in some way the most similar approaches to the problem faced in this publication: each agent defines its own plans in a parallel, distributed and independent way, until a merging algorithm is applied to merge them into an improved unique plan. This merging algorithm intends to identify key elements of the plans that provide improvements, incompatibilities, efficiencies and other relevant features in the way the problem is solved. 

This analysis and search of the key elements of plans to be merged can become pretty complex. With this goal, \cite{BruceNewman1978} proposed a structured model of actions, intentions, beliefs and states to be represented in plans. \cite{Rosenschein1982} proposed a formalization of plans to detect potential conflicts. In a similar way \cite{Georgeff1983} suggested the definition of correctness and execution conditions to be satisfied in a plan. \cite{Stuart1985} defined a temporal logic system to specify restrictions of plans that \cite{Yangetal1992} and \cite{Foulseretal1992} used to ensure an efficient merge of alternative plans that pursue a common goal.

The use of execution conditions, logic or temporal restrictions to combine plans is an approach based on the own intrinsic characteristics of the plan itself, on the dependencies of the actions that form a plan between them, which are typically domain-dependent. However, plans can also be combined using an external valuation of the actions to be included in the plan as in these two proposals:
\begin{itemize}
\item Partial Global Planning \cite{DeckerLesser1992} where each agent iteratively proposes an incremental prototype of a global plan: when an agent receives the global plan, it combines such plan with its own plan to create an improved new global plan (for instance, removing redundancies). This improvement is proposed to the other agents in such a way that they can accept, modify or reject it. 
\item Another approach based on an external estimation of the quality of a plan, comes from \cite{EphratiRosenchein1993}, where is suggested the combination of a set of plans in order to detect redundant actions through the use of A* search algorithm and a suitable costs heuristic. It was also suggested a process of joint aggregation where agents form an improved global plan voting joint actions, where votes play the role of the cost in the heuristic \cite {Ephratietal1995}.
\end{itemize}

Since these approaches are based on external evaluations of plans they are somehow similar to our proposal, but are different to our approach because the improvements are finally obtained from an internal analysis that makes use of complex knowledge representation formalisms about plans that are domain dependent and their nature is, at its core, close to the logical-temporal restrictions and execution conditions of the most extended approaches to the problem of combining plans. 

The originality of our approach relies upon the way such external evaluation is performed. We evaluate plans in a domain independent way, according to the results obtained by the past execution of plans to be combined and not to a rich (and complex) deliberation about the compatibility (and redundancies) of the actions that form the plan. It is then, an evaluation which is extrinsic of the actual composition of the plans to be combined. Roughly speaking, we could say that its underlying inspiration is closer to a neural/subsymbolic approach instead of a deliberative/symbolic: the valuation of the plan is the output of a black box with two inputs, two indirect ways of estimating the quality of the plan, as we will see below, but they are not based (as previous authors did) in the expression of the positive/negative interrelationships of the actions that form plans to be combined. 

This idea is strongly inspired on our previous works on the area of recommendation systems, specifically on the agent-based reputation management \cite{ijcis} \cite{honesty} \cite{softcomputing}. In these works, autonomous agents use the ability to dynamically combine past experiences to face new situations, selecting and mixing them to improve the expected results. 

In order to give this ability to the agents they have to represent such past experiences in a context-sensible way that includes integration mechanism based on an utility control of these experiences. Expressed in simple terms, they have to be able to distinguish (contextualize) which of these experiences are somehow (it is in fact an estimation) similar between them, and to the new situation faced. This problem has been addressed by planning systems that use case-based reasoning as in \cite{Veloso1994} \cite {Redmond1990} \cite{Goeletal1994} \cite{PlazaMcGinty2005}. Therefore solving the problem of combining past experiences (when, as in our case, such experiences are previously executed plans), requires using context information (estimation of the applicability of a particular plan to the actual situation faced by the agent, even if this estimation is roughly computed by indirect sources of information (the two inputs of the black box mentioned before).

\subsection{Hold assumptions}

Then, we were forced to accept several assumptions to apply our approach that drive us to reflect some unrealistic requirements. For instance, all the plans are been executed in a completely deterministic world. We assume that each action always produces the same consequence (no given uncertainty), and even more, it is time independent, the same results were produced any time. 

We also have to assume about the nature of autonomous agents that perform plan combinations:
\begin{itemize}
\item They are self-interested, mainly focused on reaching their own goals. 
\item They are consistent, the specification of their internal state has to be an accurate representation of the external world. 
\item Agents have no conflict between them, they are cooperative and not competitive. 
\item Agents are benevolent, they have a predisposition to help each other. 
\item Agents are honest, the information they share with others is an precise representation of their own internal state. 
\end{itemize}

In planning systems, plans are often considered as partially or completely ordered according to given restrictions over their actions, where partial ordering allows that right plans are open to different sequences of actions. On the other hand, completely ordered plans stands for an unique solution in the form of a particular sequence of actions. Our approach to the problem assumes completely ordered plans, even more, we assume that there is just one right action by each step of the plan. We also assume that actions are independent between them: the actions that are supposed to be executed before and after do not alter the suitability of a particular action ordered in the right step. 

We also assume that the final combined plan is executed by an 'operator', which can have different expertise, over a 'node' or physical element that requires with some frequency the execution of successive plans (which is dependent of the success of the previously executed plan). An action is defined by 3 values that represent its suitability:
\begin{itemize}
\item corresponding type of the node to be executed over.
\item corresponding subtype of the node to be executed over.
\item corresponding time (sequence order of the plan) to be executed in.
\end{itemize}
 
Furthermore we assume that types and subtypes of the nodes are public and previously known and that there are a limited (low) number of them. We assume that agents may compare different nodes obtaining a similarity estimation that is dependent on the type and subtype they belong to. They can then, compare the similarity of the nodes over which a plan was previously executed with the current node that requires a new combined plan to be executed. In this way, suitability of an action over a given node can be computed, it would be greater (it will reduce the expected time before the successive plan is required to be executed) when the type and subtype of the given node was similar to the type and subtype of the node that the action belongs to. 

Additionally, we assume that the number of steps a plan has is public and previously known and there are a limited (low) number of them. In this way, suitability of an action in a given plan step can be computed, it would be greater (it will reduce the expected time before the successive plan is required to be executed) when such action is placed close (or at) the right plan step. 

We assume finally that there are weights associated to the relative relevance of these three suitability criteria (type, subtype of the node and time of execution), so the final suitability of the plan is computed as a weighted sum of them. 

\section{Defining the agent system}
\begin{figure}
\centering
\includegraphics[width=9cm, height=6cm]{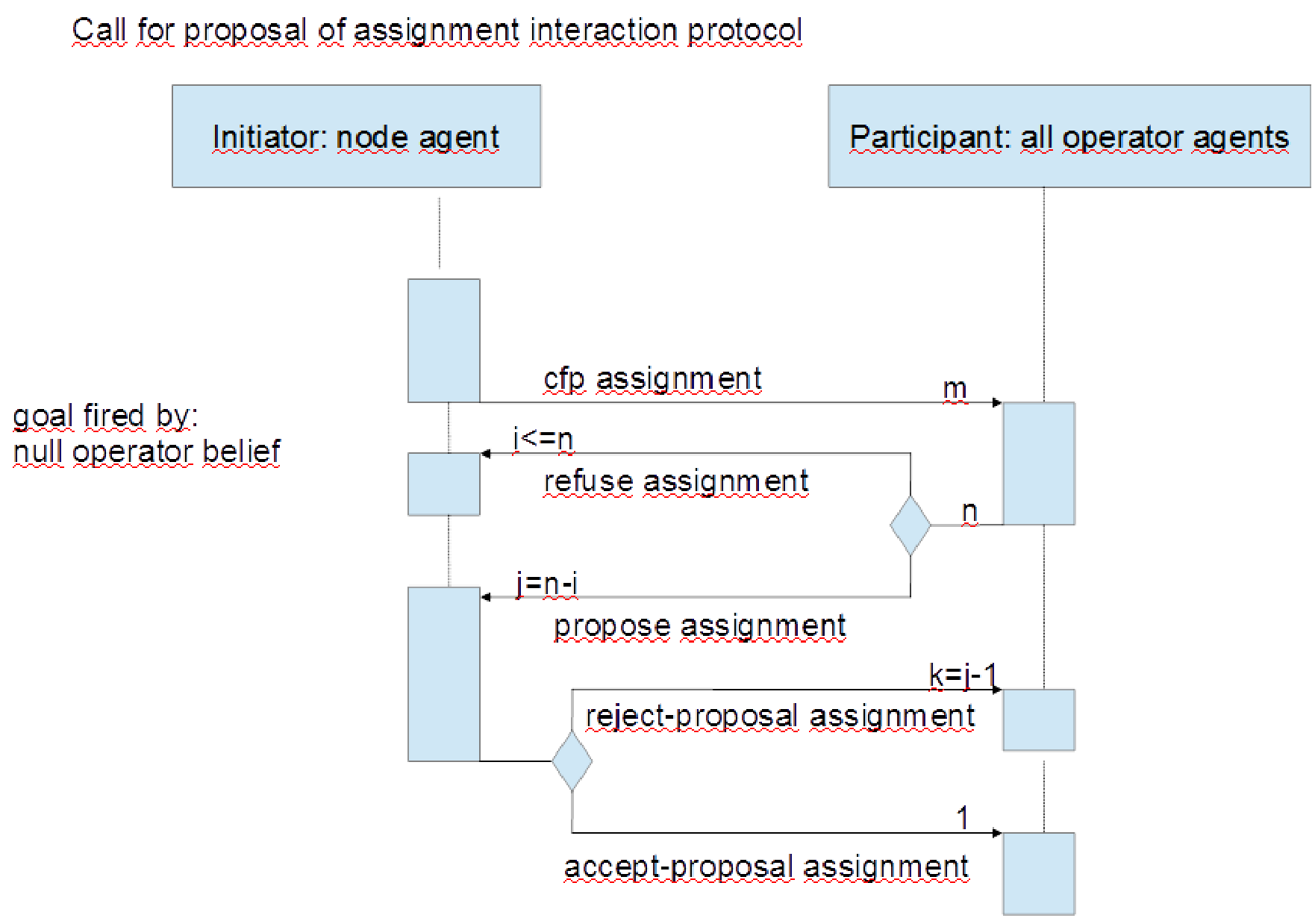}
\caption{CFP interaction protocol between a node and operator agent}
\label{fig:protocol1}
\end{figure}

\begin{figure}
\centering
\includegraphics[width=9cm, height=6cm]{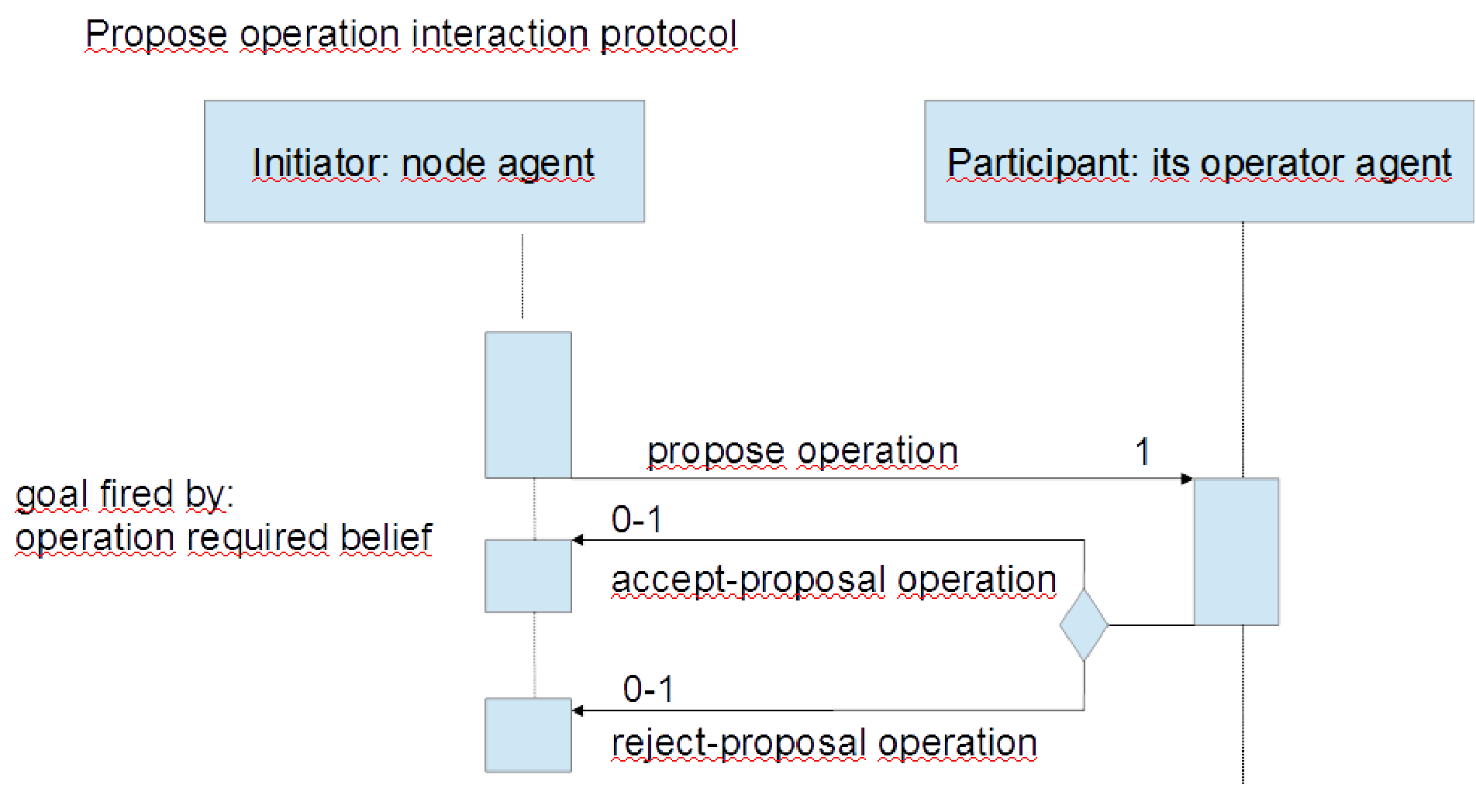}
\caption{Propose interaction protocol between a node and operator agent}
\label{fig:protocol2}
\end{figure}

\begin{figure}
\centering
\includegraphics[width=9cm, height=6cm]{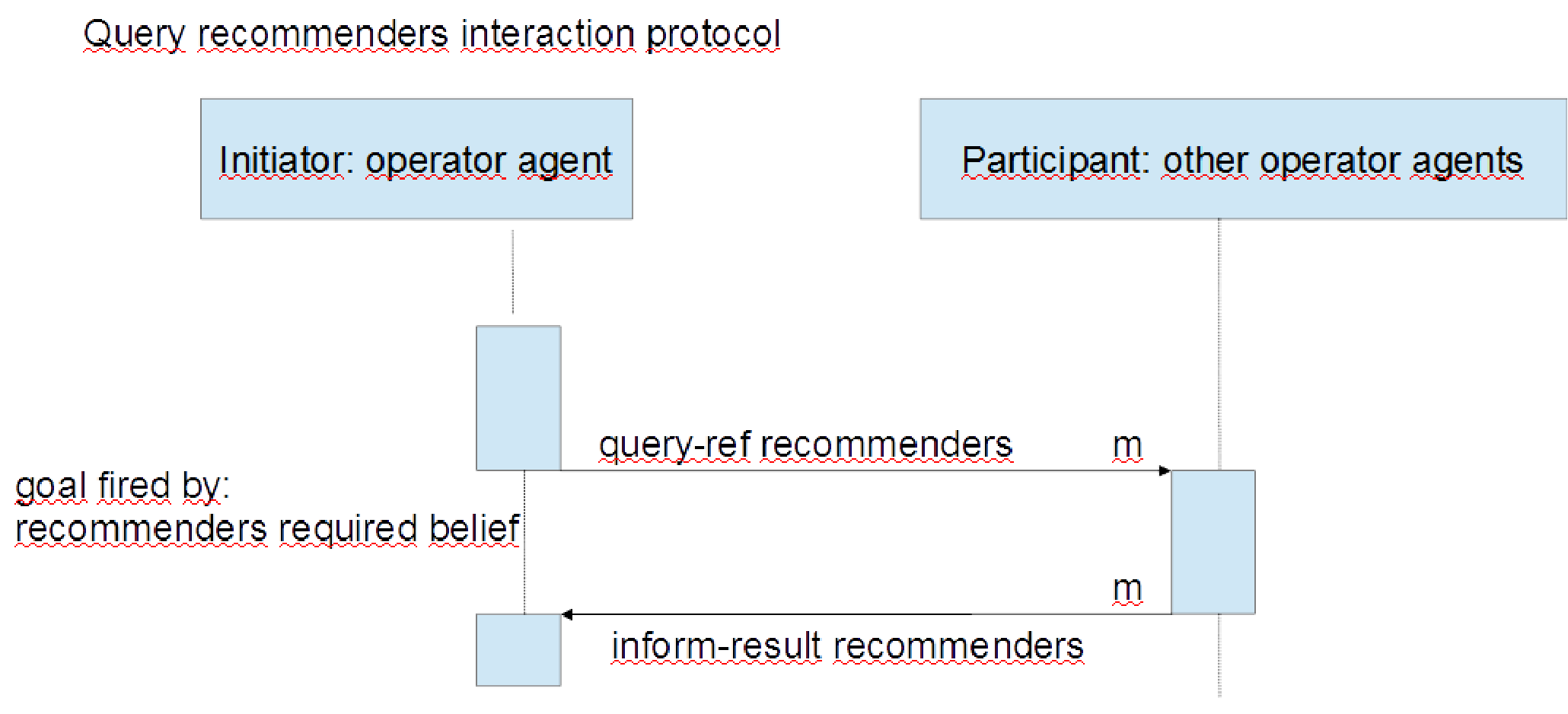}
\caption{Query recommenders interaction protocol between two operator agents}
\label{fig:protocol3}
\end{figure}

\begin{figure}
\centering
\includegraphics[width=9cm, height=6cm]{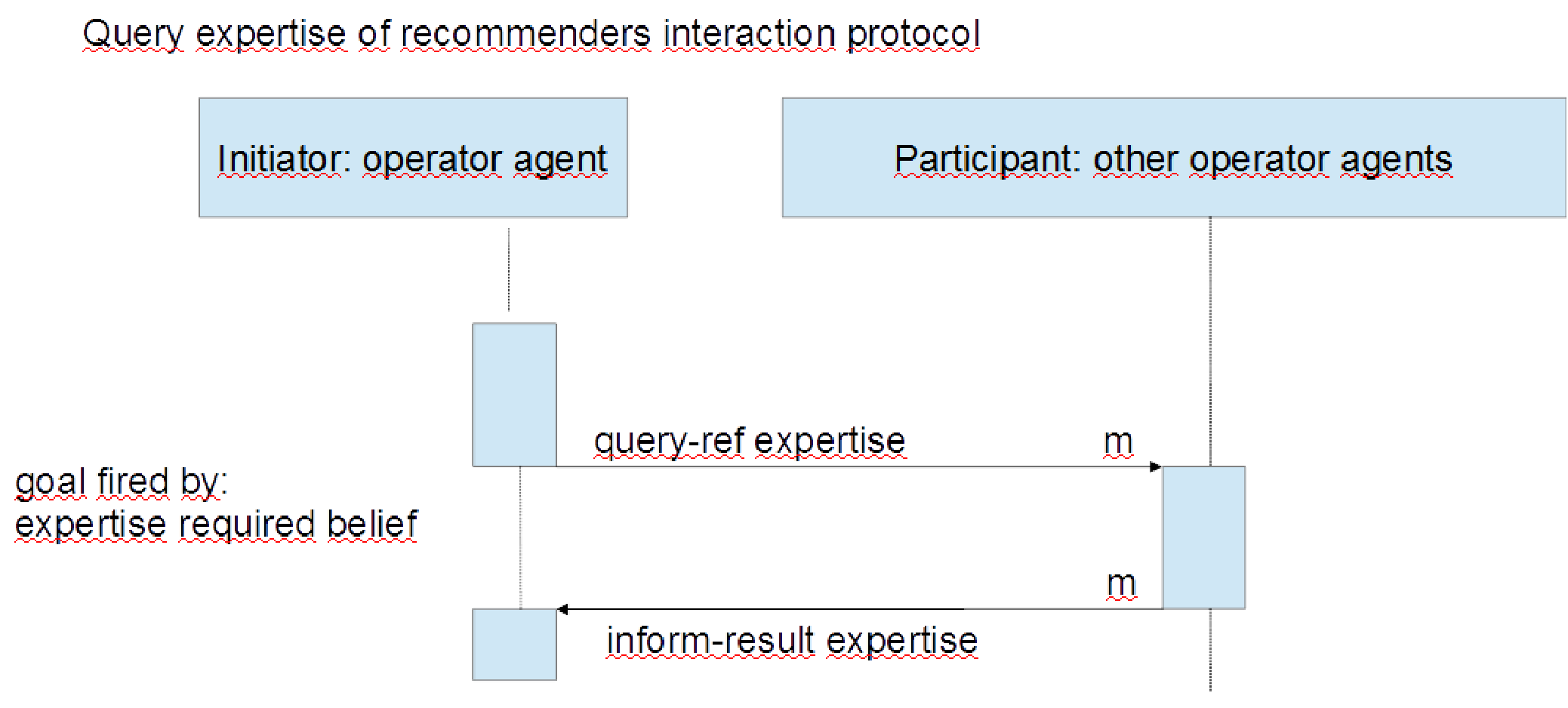}
\caption{Query expertise interaction protocol between two operator agents}
\label{fig:protocol4}
\end{figure}

\begin{figure}
\centering
\includegraphics[width=9cm, height=6cm]{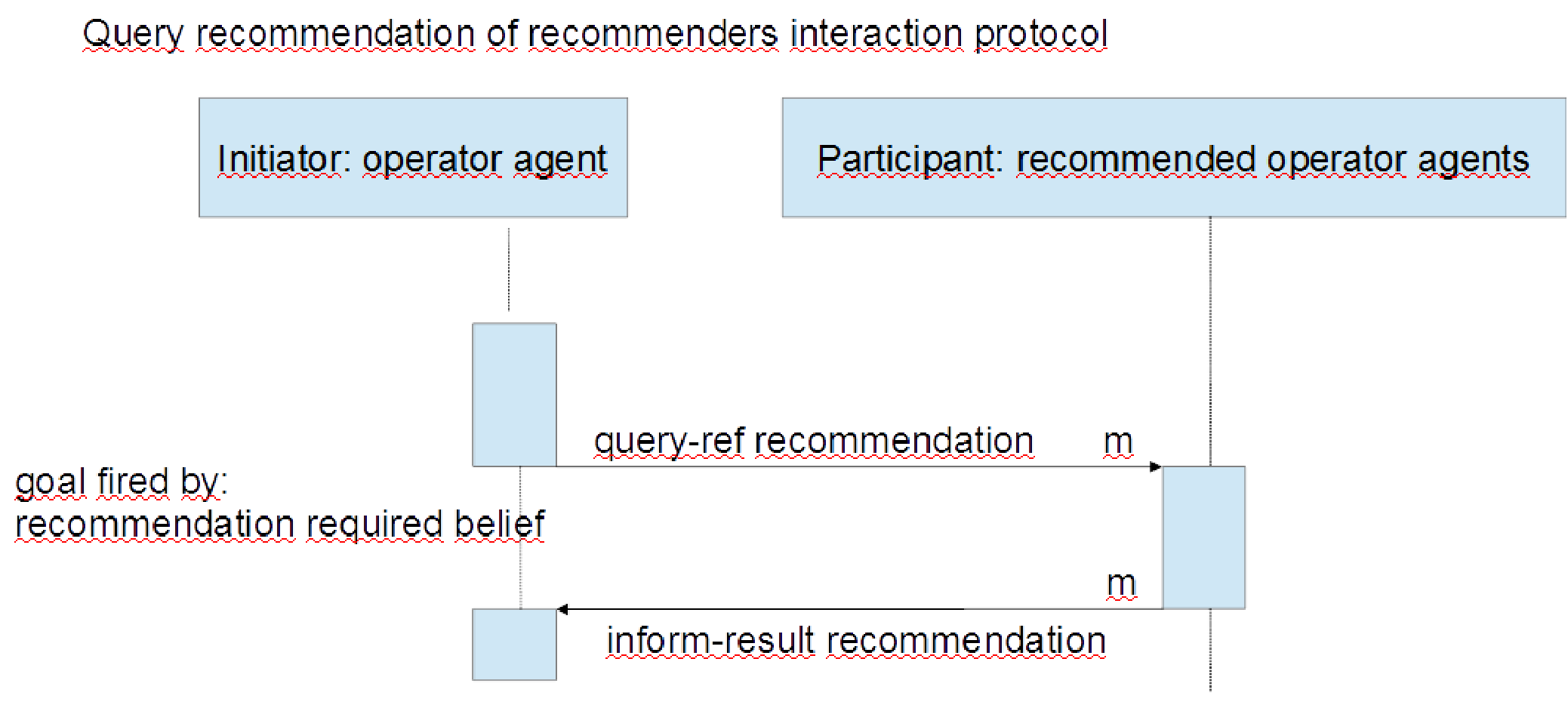}
\caption{Query recommendation interaction protocol between two operator agents}
\label{fig:protocol5}
\end{figure}

\begin{figure}
\centering
\includegraphics[width=9cm, height=6cm]{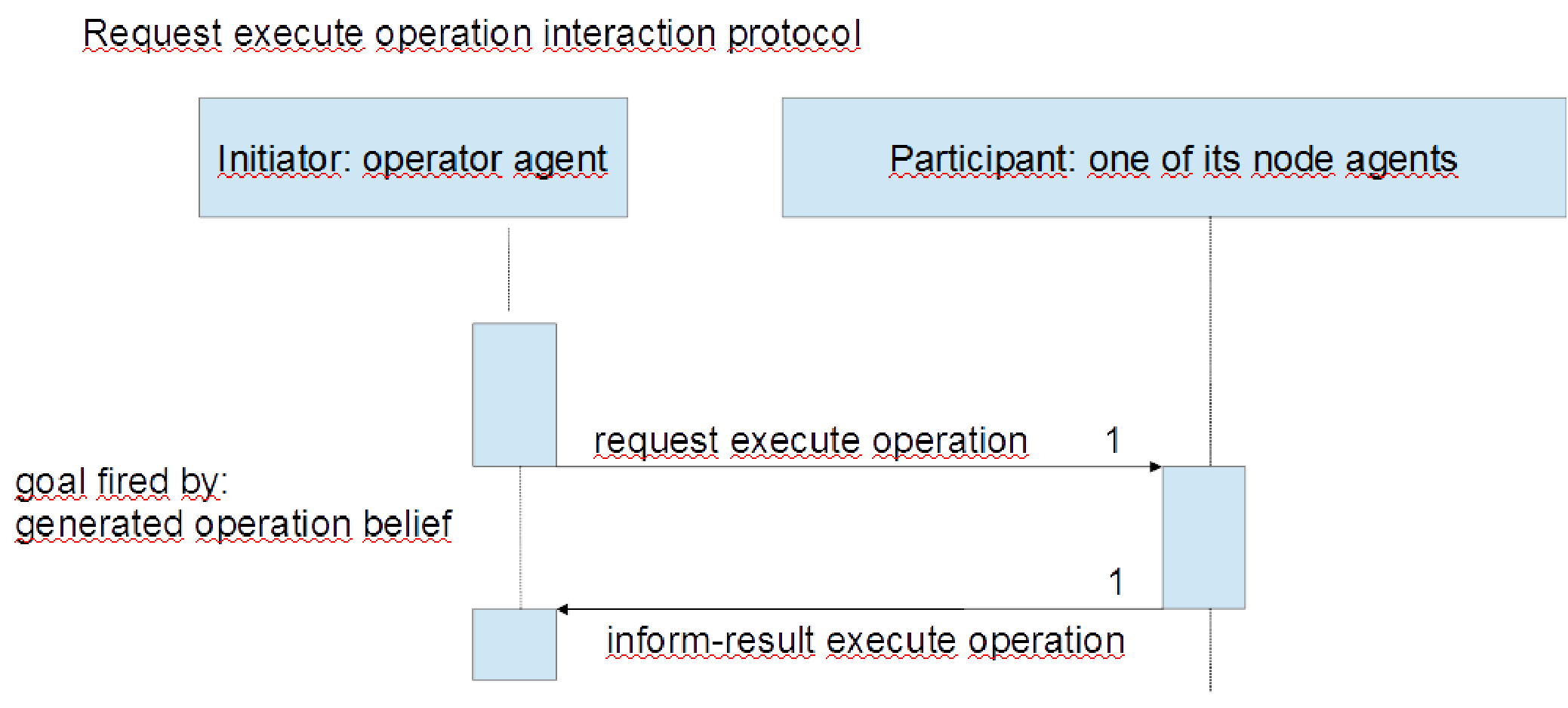}
\caption{Request plan execution interaction protocol between an operator and a node agent}
\label{fig:protocol6}
\end{figure}

Both types of agents, node and operator agents, are implemented in JADEX \cite{braubach+04jadex}, a JAVA-based academic agent platform that complies with IEEE-FIPA Agent Communication standard \cite{fipa}, and that facilitates the implementation of a deliberative approach of agent reasoning structured into three levels of knowledge: beliefs, desires and intentions. 
\begin{itemize}
\item Beliefs of agents, and the content of exchanged messages (concepts, actions and predicates of an adhoc ontology) are JAVA classes that represent such knowledge. The concepts included in the ontology are: Node identifier and type, the operation to be performed (an ordered sequence of them form a plan), which is defined by the operation type, and the node type and subtype to be applied over, expertise, which is a representation of the ability of the operator executing operation plans. On the other hand, beliefs of node agents include which is its current operator agent, the expected time until a new operation plan is required to be executed over this node, while operator agents have the next beliefs:  availability of the operator, which are the nodes currently operated, the type of the node in which is specialized, its expertise, its last operation plans performed over each type of node and finally the last operator agents that act as recommenders (identifiers and expertise) and their corresponding last recommendation.
\item Intentions of agents correspond to the actions performed by an operator or node agent looking for the satisfaction of a given goal (conceptually, a desire instantiated). Typically they involve receiving a message, accessing the content of the message and internal beliefs of the agent, perform computations over these data, and finally building and sending a response message to other node/operator agent. 
\item Desires of agents, are implemented in JADEX as goals that can be fired by external events (reception of a message) or internal events (specific conditions over beliefs of the agents are satisfied). Such goals in our agent system implementation consists of the complete execution of the corresponding role (initiator/participant in FIPA terms) in a protocol.
\end{itemize}
The messages between operator and node agents are instances of pre-defined IEEE-FIPA protocols. Therefore we will have the next sequence of protocols in a typical iteration cycle of our agent system:
\begin{itemize}
\item The first protocol to be launched is a call for proposal protocol (see figure \ref{fig:protocol1}), where node agent acts as initiator, and operator agent as participant. This protocol is associated with the goal of a node to become assigned to an operator agent, and it is fired by an internal event of a node agent because its beliefs show no current operator agent. 
\item Next, a propose interaction protocol is launched with the node agent as initiator and operator agent as participant, the goal associated is fired by an internal event when the node agent observes that the belief corresponding to the expected time until a new operation plan is required would point such circumstance out (see figure \ref{fig:protocol2}).
\item Then a query interaction protocol takes place with an operator agent acting as initiator and another one as participant of the query protocol. This query protocol pursues the identification of recommenders that are specialized in a particular type of node. It is fired by an internal event of an operator agent because its beliefs show that has to generate an operation plan for a node agent of a given type, and it has no recommenders of this node type in its beliefs (see figure \ref{fig:protocol3}). 
\item Again another query protocol is launched where an operator agent acting as initiator is asking about its expertise in a type of node to another operator acting as participant in the protocol, as a potential recommender. It is fired by an internal event of an operator agent because its beliefs show that has to generate an operation plan for a node agent, and it has to update the expertise of recommenders of this node type in its beliefs (see figure \ref{fig:protocol4}).
\item Additionally, another query protocol initiated by an operator agent ask about the operation plan that another operator agent, acting as participant in the protocol, recommends for a particular type of node. It is fired by an internal event of an operator agent because its beliefs show that has to generate an operation plan for a node agent of a given type, and it has to update the recommended operation plan of the recommenders of this node type in its beliefs (see figure \ref{fig:protocol5}).
\item Finally, a request protocol launched by the operator agent as initiator and a node agent as participant of the protocol. There, the operator agent requests the node agent to execute an operation plan which was generated through the combination of the recommendations received and the last operation plan that such operator agent executed for that type of node. It is fired by an internal event of an operator agent because its beliefs show that has to generate an operation plan for a node agent of a given type, and its beliefs corresponding to the expertise and recommended operation plans of this node type are all updated (see figure \ref{fig:protocol6}).
\end{itemize}

\section{Merging Plans}

\subsection{Computing the reputation of an operator agent and the required time for the next plan execution}
Therefore, the combination of plans would take place before the execution of this last protocol when the operator agent has the next updated knowledge in its beliefs:
\begin{itemize}
\item expertise of operator agents in the node type acting as recommenders
\item operation plan recommended by operator agents for the given node type
\item previous own expertise in the node type
\item previous own operation plan in the node type
\end{itemize} 
These four inputs would generate a new operation plan that would be sent to the node type, and its corresponding answer would include the time to be spent before a new operation plan would be required (computed according to the level of success obtained with the operation plan suggested by the operator agent), which will be used to update the own expertise of the operator agent in such node type.

We will obtain this value of the level of success inspired by the equations that the Agent Reputation Testbed (ART) platform \cite{art} used. Such platform have been used for experimentation in several publications \cite{iam} \cite{simplet} \cite{peles} \cite{uno2008} and for several international competitions hold in International Conference on Autonomous Agents and Multi Agent Systems (AAMAS).

This platform generates the error in a service estimation through a normal distribution centered in the difference between the estimated (noted as e) and the true value of the service (noted as t), with a standard deviation s that stands for the noise that avoids an optimal perception of the true value of the service in real life (equation \ref{eq:art0}). s takes its value from equation \ref{eq:art}. In this equation, E stands for the expertise of the agent in providing such service, c stands for the cost for the agent in providing the service, in order words, the time (the effort) invested by the agent. In this way the bigger the expertise, the invested effort, and the success of the estimation the less noise would be applied to the normal distribution.   

\begin{equation}\label{eq:art0}
    resul = \Phi ( \left\| e - t \right\| , s )
\end{equation}

\begin{equation}\label{eq:art}
    s=( 1 - E + 1/c ) \times \left\| e - t \right\|  
\end{equation}

In the same way, may compute the difference between the optimal plan and the suggested plan (node agents compute such difference). In order to do it, we assume the existence of such optimal plan, and that plans have a limited number of steps, and that the possible operations to be applied in each step is also limited and dependent of the type and subtype of the nodes. Therefore, the number of steps, the number of types and subtypes, and the weights of making a mistake in the step, the type and the subtype are all setup parameters of the agent system.

Then, each operation of an optimal plan is defined by three dimensions: the right type of the node to be applied in, the right subtype of the node and the right time step to be applied when. The difference of a operation belonging to the plan suggested by an operator agent with the optimal one is computed as a weighted sum of the distance of the node type, subtype and timestep (sequence order) as it is shown in equation \ref{eq:dif}. 

\begin{equation}\label{eq:dif}
\begin{split}
   \left\| e - t \right\| = \sum_{i=1-maxtimestep} ( w_{type} \times \left\| e_{i,type} - t_{i,type} \right\| \\
 %&\quad \cap
	+ w_{subtype} \times\left\| e_{i,subtype} - t_{i,subtype} \right\| \\
 + w_{timestep} \times \left\| e_{i,timestep} - t_{i,timestep} \right\| )
	\end{split}  
\end{equation}

Once we have the level of success obtained with the plan suggested by the operator agent, the node agent has to compute the the required time for the next plan execution. It would take its value from equation \ref{eq:time}, where resul stands for the level of success of the suggested plan and maxtime stands for the maximum time that any node may spend without the execution of a new plan (given static value as paremeter in the setup of the agent system). In this way the better the level of success of the plan, the less time would be required for a new plan to be executed.    

\begin{equation}\label{eq:time}
    time = resul \times maxtime 
\end{equation}

Finally the node agent has to compute the new expertise of the operator agent after the execution of the suggested plan. In our design, the concept of expertise has two associate values:
\begin{itemize}
\item Global expertise stands for the number of times an operator agent has executed plans (whatever their success) in a given node type. It will increase until a threshold (defined as a setup parameter), and after that number of plan executions, global expertise would be fixed at the maximum value. Global expertise would be used instead of the invested time of equation \ref{eq:art}, assuming that our operator agents spend all of them the same amount of time in the execution of plans, but operator agents with more global expertise would be more efficient in the same time.
\item Local expertise stands for the ability just showed in suggesting a plan for that node type. It is computed as the opposite of the standard deviation s, and then it has an recursive definition, past local expertise influences the computation of the current local expertise, as s was defined in equation \ref{eq:artmod} as dependent of local expertise. 
\end{itemize}

According to these two features of expertise, equation \ref{eq:art} from ART testbed becomes redefined in our agent system as equation \ref{eq:artmod}.

\begin{equation}\label{eq:artmod}
    s=( 1 - E_l + 1/E_g ) \times \left\| e - t \right\|  
\end{equation}

On the other hand operator agents have to weight the recommended plans from other operator agents in order to combine them into a new single plan to suggest to the node agent. Reputation of operator agents in our domain is contextual, in other words, it would take a different value for each node type. It will be computed from the (global and local) expertise that the operator agent has using equation \ref{eq:reputation}. Where $E_l$ stands for local expertise, $E_g$ stands for global expertise, and Thr stands for the threshold of global expertise. 

\begin{equation}\label{eq:reputation}
    rep = 1 - ( 1 - E_l ) \times ( E_g / Thr )
\end{equation}

Next we will show how all these elements may be used to combine plans in 4 different ways but the first of them is just a no-merging method used as a benchmark to compare how much combining plans improve the level of success of operator agents suggesting a plan to the node agent.

\subsection{Merging method 0: no merging ignoring recommendations}

The operator agent would ignore all received recommendations, and it would execute plans updated exclusively from its own previous execution of plans as it is shown in equation \ref{eq:merge0} where s stands for the success of the execution of previous plans. This method is used as benchmark, and the relative benefits of applying the other methods can therefore be measured by the improvements in results obtained with respect the results of this no-cooperation merging method.

\begin{equation}\label{eq:merge0}
    new plan = oldplan_j , j = \max_{i=1-n} s_{i}
\end{equation}

\subsection{Merging method 1: no merging considering recommendations}

The operator agent would choose the best plan (according to the computed reputation of operator agents) among all the possible options (recommendations and own previous plan for that node type) as a whole, with no combination at all  as it is shown in equation \ref{eq:merge1} where rep stands for the reputation of the received recommendation. 

\begin{equation}\label{eq:merge1}
    new plan = recommendedplan_j , j = \max_{i=1-n} rep_{i}
\end{equation}

\subsection{Merging method 2}

The operator agent would build the new plan combining the operations that have more reputation (inherited by the operator agent that suggested this operation in this plan step in its recommended plan) and popularity (the number of times this operation is suggested by operator agents). In other words, we sum all the reputations of each possible operation for each plan step, so repeated operations and operations from operators with good reputation have more options to become the operation suggested in the new plan step. This process is summarized in equation \ref{eq:merge2} where numoper stands for the number of different recommended operations for a given time step j and numrec stands for the number of recommenders that suggested such operation i to be applied in a given time step j. 

\begin{equation}\label{eq:merge2}
    new plan = [_{k=1-numtimesteps}  \max_{i=1-numoper_k}  \sum_{j=1-numrec_{k,i}} rep_{i,j} ]
\end{equation}

\subsection{Merging method 3}

The operator agent modifies the recommended plan from the operator agent with most reputation. The modification consists in replacing a number of the operations that this best plan (according to the reputation) shares in the same plan step with the worst plan. These operations would be replaced by different operations suggested by the other recommended plans (chosen in decreasing order of reputation of the operator agent). The number of them is fixed as a parameter setup. Therefore, this combination method is someway similar to evolutive algorithms, where plans are the individuals to be crossed and reputation plays the role of fitness function. This process is summarized in equation \ref{eq:merge3} where i goes from 1 to the number of operations recommended for timestep k minus one, which is the operation that the best and the worst plans share.

\begin{equation}\label{eq:merge3}
\begin{split}
    best plan = recommendedplan_j , j = \max_{i=1-n} rep_{i} \\
    worst plan = recommendedplan_j , j = \min_{i=1-n} rep_{i} \\
		new plan = [_{k=1-numtimesteps}  best_k \neq worst_k \rightarrow best_k \\
		else \max_{i=1-(numoper_k-1)}  \sum_{j=1-numrec_{k,i}} rep_{i,j} ]
\end{split}
\end{equation}

\section{Experimentation}
\begin{figure}
\centering
\includegraphics[width=9cm, height=6cm]{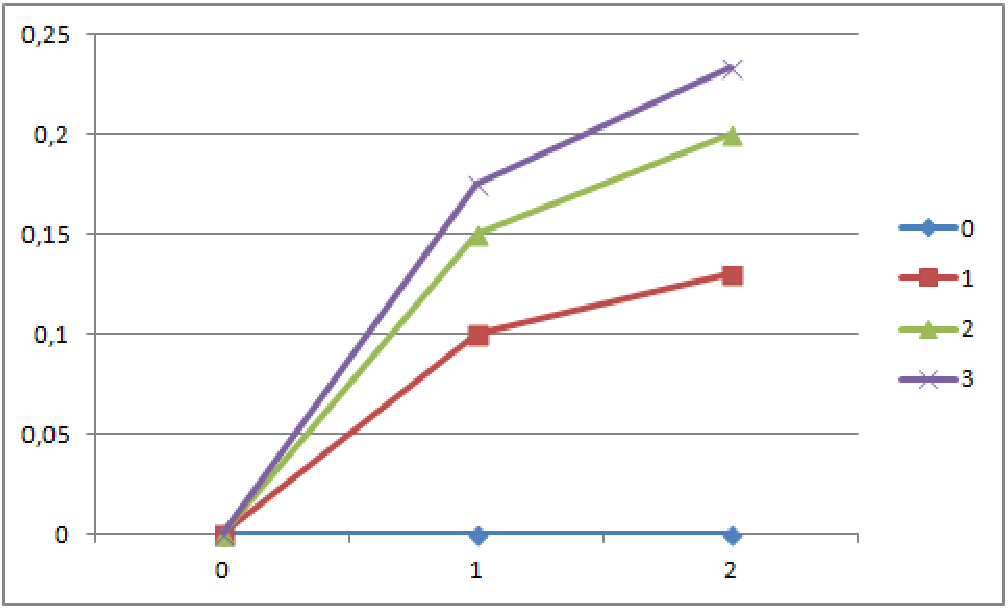}
\caption{Improvement due to merging methods with 20 operators and 1 node along 2 iterations}
\label{fig:1agent2iter}
\end{figure}

We have carried out several simulations to test the efficiency of the four merging methods described above. First, we have run a very simple simulation with next parameters setup: 
\begin{itemize}
\item Nodes have just a type and two subtype.
\item each plan is composed of 5 steps.
\item each operator takes charge of a single node (availability equal to 1).
\item each operator may ask upto 20 recommenders.
\item each operator has an initial expertise equal to 1.
\item the weights associated to the errors produced mispredicting node type, subtype and plan operation are all equal to 1. 
\item 20 operator agents.
\item number of replacements of merging method 3 equal to 1.
\item 1 node agent
\item the previous plans of the operator agents are randomly generated with an uniform distribution.
\item The node agent would choose one out 20 operator agents, and such chosen operator agent would ask for recommendations to the other 19 operators about its previous experience, the resulting plan would combine the 20 randomly generated experiences. As there is no more node agents, and the other 19 operator agents just act as recommenders (with a constant recommendation), it has no sense to run more than 2 iterations. 
\end{itemize}
Such simplistic simulation run 4 times (one for each merging method), gave the results of \ref{fig:1agent2iter}.  
The vertical axis of figure \ref{fig:1agent2iter} shows how the combined plan of each merging method is improving (is closer to the optimal plan) while the horizontal axis shows the evolution along the 2 iterations. The resulting lines of each merging methods seem to confirm the order in which they obtain the best possible plan from the same available information through cooperation (in form of recommendations) but without any evolution of the available information of recommendations. 

\begin{figure}
\centering
\includegraphics[width=9cm, height=6cm]{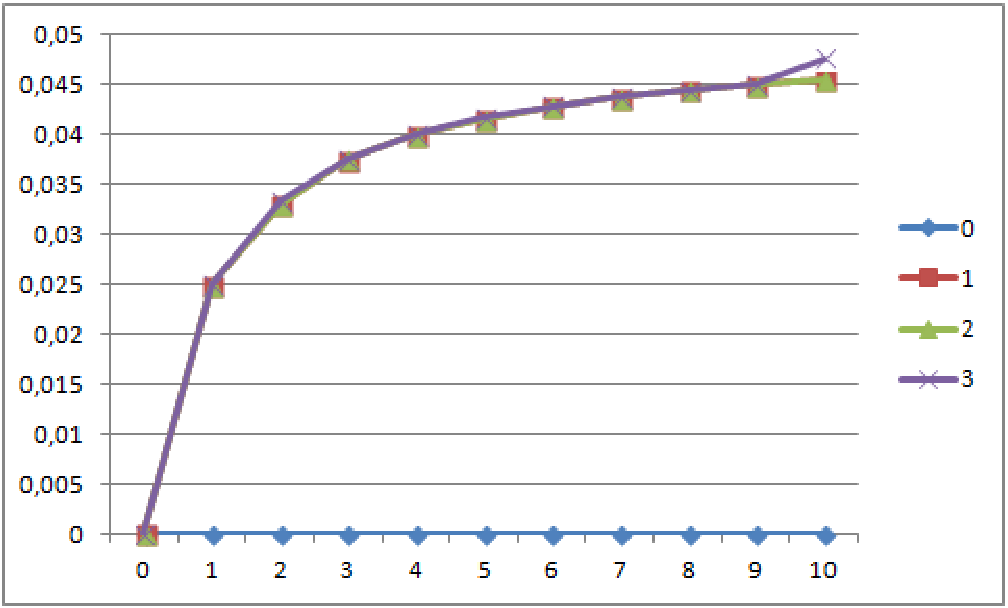}
\caption{Improvement due to merging methods with 2 operators and 2 nodes along 10 iterations}
\label{fig:2agent10iter}
\end{figure}

A second set of 4 simulations, figure \ref{fig:2agent10iter},  decrease the level of cooperation (2 operator agent), but now information provided by recommendations is improving in each iteration (2 node agents and 10 iterations). Each operator agent would be linked to a node agent (their availability remains equal to 1), they would be sharing their relative success and failures along the 10 iterations. 
The figure \ref{fig:2agent10iter} shows the same information in both axis, and the lines shows how evolves the improvement reached (in average) of both operator agents. Here we can observe that the relative improvements are much less significant ($0.05<<0.25$)when operator agents combine its plan with just one recommendation (instead of 19 in the previous simulation). So merging methods act much better when they have less plans to combine, even when they are not accurate or updated (as they were constant in the previous simulation). Additionally the differences between 'serious' merging methods (1-3) are very small in these circumstances. 

\begin{figure}
\centering
\includegraphics[width=9cm, height=6cm]{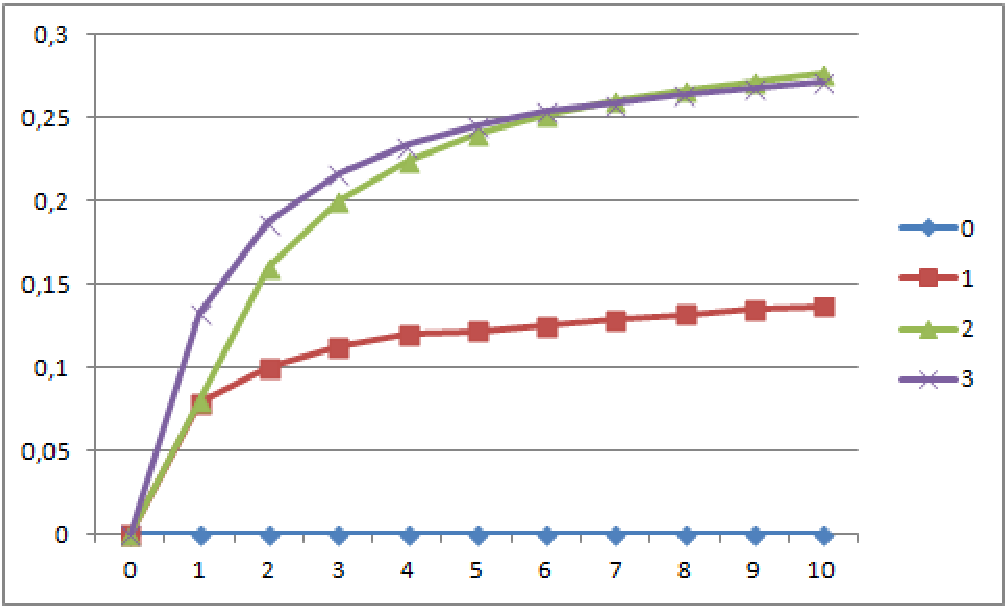}
\caption{Improvement due to merging methods with 10 operators and 10 nodes along 10 iterations}
\label{fig:10agent10iter}
\end{figure}

So at last, we run a final set of 4 simulations, figure \ref{fig:10agent10iter}, this time with 10 node agents, 10 operator agents and along 10 iterations, in order to see how both features (evolution due to changes in previous experiences, and cooperation through accurate recommendations). 
In figure \ref{fig:10agent10iter} we observe now how the differences between merging methods that first simulation showed now arose again, and that the relevance of the improvement in the combined plan generated in each iteration has a similar scale ($0.3\approx 0.25$). Merging method 1 shows to be generating combined plans not so close to the optimal one, while merging methods 2 and 3 shows similar results, although merging method 3 seems to show a slight better results in the first iterations (faster convergence). 
From the view of the simulations hold, we can roughly estimate that most of the improvement is due to cooperation of agents in form of recommendations (method 1 is close to methods 2 and 3) instead of the improvement from combination of plans. Therefore, it seems that, although any combination method to merge plans is required to take advantage of recommendations, most improvement would be expected from the research on the issues from recommendation systems domain (computing reputation of sources and selecting the right partners to cooperate with) than from the issues from planning domain relative to merging plans.  
Finally, we also know that the set of hold simulations of our agent system is limited, and many other different initial setups could be considered. But our thoughts were driven to the idea of keeping it simple to observe with more clarity the relative contribution of merging plans and the use of recommendations.  
\section {Conclusions}
In this paper we have accomplished the goals we pursued:
\begin{itemize}
\item we have defined and implemented an agent system that allows the comparison of alternative ways to merge plans through the roles of nodes and operators using BDI paradigm.
\item we have assumed a given number of conditions (mainly that actions are independent between them) that provide the possibility of an evaluation of the actions to be included in the plan that do not depend on intrinsic domain-dependent restrictions over the actions. 
\item we have justified the use of recommendations and defined a way to weight them according to the past execution of plans, and several (simple) ways to merge them.
\item we have tested with specific setups the performance of the merging algorithms defined and observed that the contribution from using recommendations is much greater than from the way plans are merged.
\end{itemize}
Our contribution is innovative since it is located in the boundaries of planning and recommender systems research issues, relevant since its applicability may be not restricted to domain dependencies logic, and it provides new paths to be explored by other researchers since our agent system is an open (and available in sourceforge \footnote{webtobeincluded}) framework ready to include many other merging algorithms, different ways of computing reputation of recommenders, or just test with any other initial setup. 

This paper also provides an useful tool to help people with cognitive disabilities such as autism managing transition planning since our proposal pursues a domain-independent completely autonomous way to generate such plans.

%\end{abstract}
%\label{}



\bibliographystyle{elsarticle-num} 
\bibliography{bibs}

\begin{thebibliography}{10}
\expandafter\ifx\csname url\endcsname\relax
  \def\url#1{\texttt{#1}}\fi
\expandafter\ifx\csname urlprefix\endcsname\relax\def\urlprefix{URL }\fi
\expandafter\ifx\csname href\endcsname\relax
  \def\href#1#2{#2} \def\path#1{#1}\fi

\bibitem{ShogrenPlotner2012}
K.~A. Shogren, A.~J. Plotner, Transition planning for students with
  intellectual disability, autism, or other disabilities: Data from the
  national longitudinal transition study, Intellectual and Developmental
  Disabilities 50~(1) (2012).

\bibitem{Ronmark2014}
L.~Ronmark, The Never Ending Shower : planning ability, intellectual disability
  and cognitive artifacts, Master Thesis, Linkoping University, Department of
  Computer and Information Science, 2014.

\bibitem{EphratiRosenchein1993}
E.~Ephrati, J.~S. Rosenschein, Multi-agent planning as a dynamic search for
  social consensus, in: Proceedings of the Thirteenth International Joint
  Conference on Artificial Intelligence, Chambery, France, 1993, pp. 423--429.

\bibitem{Rosenschein1982}
J.~S. Rosenschein, Synchronization of multi-agent plans, in: Proceedings of the
  National Conference on Artificial Intelligence, Pittsburgh, Pennsylvania,
  1982, pp. 115--119.

\bibitem{Georgeff1983}
M.~P. Georgeff, Communication and interaction in multi-agent planning., in:
  M.~R. Genesereth (Ed.), AAAI, AAAI Press, 1983, pp. 125--129.

\bibitem{MuscettolaSmith1987}
N.~Muscettola, S.~F. Smith, A probabilistic framework for resource-constrained
  multi-agent planning, in: Proceedings of the 10th International Joint
  Conference on Artificial Intelligence - Volume 2, IJCAI'87, Morgan Kaufmann
  Publishers Inc., San Francisco, CA, USA, 1987, pp. 1063--1066.

\bibitem{WilkinsMyers1998}
D.~E. Wilkins, K.~L. Myers, A multiagent planning architecture, in: Proceedings
  of the Fourth International Conference on Artificial Intelligence Planning
  Systems, Pittsburgh, Pennsylvania, USA, 1998, 1998, pp. 154--162.

\bibitem{Fischeretal1995}
K.~Fischer, J.~P. M{\"{u}}ller, M.~Pischel, D.~Schier, A model for cooperative
  transportation scheduling, in: Proceedings of the First International
  Conference on Multiagent Systems, June 12-14, 1995, San Francisco,
  California, {USA}, 1995, pp. 109--116.

\bibitem{BruceNewman1978}
B.~C. Bruce, D.~Newman, Interacting plans, Cognitive Science~(2) (1978)
  195--233.

\bibitem{Stuart1985}
S.~M. Shieber, Using restriction to extend parsing algorithms for
  complex-feature-based formalisms, in: 23rd Annual Meeting of the Association
  for Computational Linguistics, 8-12 July 1985, University of Chicago,
  Chicago, Illinois, USA, Proceedings., 1985, pp. 145--152.

\bibitem{Yangetal1992}
Q.~Yang, D.~S. Nau, J.~A. Hendler, Merging separately generated plans with
  restricted interactions., Computational Intelligence 8 (1992) 648--676.

\bibitem{Foulseretal1992}
D.~E. Foulser, M.~Li, Q.~Yang, Theory and algorithms for plan merging,
  Artificial Intelligence Journal 57~(2-3) (1992) 143--182.

\bibitem{DeckerLesser1992}
K.~S. Decker, V.~R. Lesser, Generalizing the partial global planning algorithm,
  International Journal of Intelligent and Cooperative Information Systems 1
  (1992) 319--346.

\bibitem{Ephratietal1995}
E.~Ephrati, M.~E. Pollack, J.~S. Rosenschein, A tractable heuristic that
  maximizes global utility through local plan combination, in: Proceedings of
  the First International Conference on Multiagent Systems, June 12-14, 1995,
  San Francisco, California, {USA}, 1995, pp. 94--101.

\bibitem{ijcis}
J.~Carbo, J.~Molina, J.~Davila, Trust management through fuzzy reputation,
  International Journal of Cooperative Information Systems 12~(1) (2003)
  135--155.

\bibitem{honesty}
M.~Gomez, J.~Carbo, C.~Benac, Honesty and trust revisited: the advantages of
  being neutral about other's cognitive models, Journal Autonomous Agents and
  Multi-Agent Systems (JAAMAS) 15~(3) (2007) 313--335.

\bibitem{softcomputing}
J.~Carbo, J.~M. Molina, An extension of a fuzzy reputation agent trust model
  (afras) in the art testbed, Soft Computing 14~(8) (2010) 821--831.

\bibitem{Veloso1994}
M.~M. Veloso, Planning and Learning by Analogical Reasoning, Vol. 886 of
  Lecture Notes in Computer Science, Springer, 1994.

\bibitem{Redmond1990}
M.~Redmond, Distributed cases for case-based reasoning: Facilitating use of
  multiple cases, in: Proceedings of the 8th National Conference on Artificial
  Intelligence. Boston, Massachusetts, July 29 - August 3, 1990, 2 Volumes.,
  1990, pp. 304--309.

\bibitem{Goeletal1994}
A.~K. Goel, K.~S. Ail, M.~W. Donnellan, A.~G. de~Silva~Garza, T.~J. Callantine,
  Multistrategy adaptive path planning, {IEEE} Expert 9~(6) (1994) 57--65.

\bibitem{PlazaMcGinty2005}
E.~PLAZA, L.~MCGINTY, Distributed case-based reasoning, The Knowledge
  Engineering Review 20~(03) (2005) 261--265.

\bibitem{braubach+04jadex}
L.~Braubach, A.~Pokahr, W.~Lamersdorf, Jadex: A short overview, in: Main
  Conference Net.ObjectDays 2004, 2004, pp. 195--207.

\bibitem{fipa}
Foundations for intelligent phisical agents specification, Geneve, Switzerland,
  1997.

\bibitem{art}
K.~Fullam, T.~Klos, G.~Muller, J.~Sabater, A.~Schlosser, Z.~Topol, K.~S.
  Barber, J.~Rosenschein, L.~Vercouter, M.~Voss, A specification of the agent
  reputation and trust (art) testbed: Experimentation and competition for trust
  in agent societies, in: The Fourth International Joint Conference on
  Autonomous Agents and Multiagent Systems (AAMAS-2005), 2005, pp. 512--518.

\bibitem{iam}
W.~LukeTeacy, T.~Huynh, R.~Dash, N.~Jennings, J.~Patel, M.~Luck, The art of
  iam: The winning strategy for the 2006 competition, in: Procs. of Trust in
  Agent Societies WS Procs., AAMAS 2007, 2007.

\bibitem{simplet}
J.~F.~H. Yann~Krupa, L.~Vercouter, {Extending the Comparison Efficiency of the
  ART Testbed}, in: M.~Paolucci (Ed.), Proceedings of the First International
  Conference on Reputation: Theory and Technology - ICORE 09, Gargonza, Italy,
  2009.

\bibitem{peles}
A.~Diniz Da~Costa, C.~J. Lucena, V.~Torres Da~Silva, S.~C. Azevedo, F.~A.
  Soares, Trust in agent societies, Springer-Verlag, Berlin, Heidelberg, 2008,
  Ch. Art Competition: Agent Designs to Handle Negotiation Challenges, pp.
  244--272.

\bibitem{uno2008}
V.~Munoz, J.~Murillo, B.~Lopez, D.~Busquets, Strategies for exploiting trust
  models in competitive multi-agent systems, in: L.~Braubach, W.~van~der Hoek,
  P.~Petta, A.~Pokahr (Eds.), Multiagent System Technologies, Vol. 5774 of
  Lecture Notes in Computer Science, Springer Berlin / Heidelberg, 2009, pp.
  79--90.

\end{thebibliography}


%\begin{thebibliography}{00}


%\bibitem{}

%\end{thebibliography}
\end{document}